\documentclass{article} 
\usepackage{iclr2023_conference,times}
\pdfoutput=1

\usepackage{amsmath,amsfonts,bm}

\def\eqref#1{equation~\ref{#1}}

\def\1{\bm{1}}

\DeclareMathAlphabet{\mathsfit}{\encodingdefault}{\sfdefault}{m}{sl}
\SetMathAlphabet{\mathsfit}{bold}{\encodingdefault}{\sfdefault}{bx}{n}

\usepackage[pdftex]{hyperref}
\hypersetup{
	pdfborder={0 0 0},
	colorlinks,
	citecolor={[rgb]{0.0, 0.0, 0.5}},
	linkcolor={[rgb]{0.0, 0.0, 0.5}},
	urlcolor=black
}
\usepackage{graphicx}

\usepackage{wrapfig}
\usepackage{caption}
\usepackage{subcaption}
\usepackage{algorithm}
\usepackage[noend]{algpseudocode}

\title{Meta-learning for unsupervised outlier detection with optimal transport}

\author{Prabhant Singh \& Joaquin Vanschoren  \\
Eindhoven University of Technology\\
Eindhoven 5600 MB, Netherlands \\
\texttt{\{p.singh,j.vanschoren\}@tue.nl} \\
}

\iclrfinalcopy 

\begin{document}

\maketitle
\begin{abstract}
Automated machine learning has been widely researched and adopted in the field of supervised classification and regression, but progress in unsupervised settings has been limited. We propose a novel approach to automate outlier detection based on meta-learning from previous datasets with outliers. Our premise is that the selection of the optimal outlier detection technique depends on the inherent properties of the data distribution. We leverage optimal transport in particular, to find the dataset with the most similar underlying distribution, and then apply the outlier detection techniques that proved to work best for that data distribution. We evaluate the robustness of our approach and find that it outperforms the state of the art methods in unsupervised outlier detection. This approach can also be easily generalized to automate other unsupervised settings.
\end{abstract}

\section{Introduction}
Outlier detection(OD) is the process of identifying data points that are significantly different from the rest of the data. These data points can be caused by errors in the data collection process, incorrect values, or unusual events. Outlier detection can be used to improve the quality of the data or to find unusual events that could be interesting to different business and scientific domains . The term "outlier detection" can be interchangeably used with "anomaly detection". For consistency, we will use the term "outlier detection" in this paper. Outlier detection has multiple applications such as medicine ~\citep{Chow1990OutlierDI, Ma2021OutlierDI}, chemistry ~\citep{Egan1998OutlierDI} and molecular biology ~\citep{Cho2016OutlierDF}. Outlier detection has been a particularly hard problem. A number of Outlier detection algorithms have been introduced in the last two decades ~\citep{Aggarwal2013OutlierA}. Unsupervised outlier detection is a very challenging task with no universally good model which works optimally on every task ~\citep{Campos2015OnTE}.
\par
AutoML ~\citep{Hutter2019AutomatedML} has shown reliable performance and benefits in model selection and hyperparameter optimization ~\citep{Hutter2019AutomatedML,NIPS2015_11d0e628, Thornton2013}. The research in Automated machine learning has been highly focused on supervised machine learning where we can focus on the performance on the hold-out dataset to define an optimization metric for the search algorithm which finds the optimal algorithms by iterating over the search space. This setting is very reliable ~\citep{NIPS2015_11d0e628} but the research on unsupervised setting is rather limited. In recent years frameworks like MetaOD ~\citep{metaod} have appeared which attempt to solve automated outlier detection via meta-learning ~\citep{Vanschoren2018MetaLearningAS}. 
\par

In this work we propose an automated framework for unsupervised machine learning tasks \textbf{LOTUS}(Learning to learn with Optimal Transport for Unsupervised Scenarios), which leverages meta-learning ~\citep{Vanschoren2018MetaLearningAS} and optimal transport distances ~\citep{Peyr2019ComputationalOT, Scetbon2022LowrankOT}. In this work we use LOTUS to perform model selection on a given unsupervised outlier detection task. We also develop \textbf{GAMAOD}, an extension to the popular AutoML framework GAMA ~\citep{gama} for supervised outlier detection problems. 
In summary, we make the following 4 contributions:
\begin{itemize}
    \item \textbf{A Meta-learner for outlier detection}: We propose \textbf{LOTUS}: Learning to learn with Optimal Transport for Unsupervised Scenarios, an optimal transport based meta-learner which recommends an optimal outlier detection algorithm based on a historical collection of datasets and models in a zero-shot learning scenario. Our solution can be used in cold start settings for model selection on unsupervised outlier detection.
    \item\textbf{ A supervised automated framework} for unsupervised outlier detection: We develop \textbf{GAMAOD} which is an extension of an AutoML tool GAMA. GAMAOD can use different search algorithms to find optimal unsupervised outlier detection algorithms for a given dataset when labels are available. We develop GAMAOD to populate the meta-data for model selection. 
    \item \textbf{Experiments and results:} We empirically evaluate LOTUS in combination with existing state of the art methods. We demonstrate the robustness of our approach against existing state of the art meta-learners and learners.
    \item \textbf{Open source:} We open-source the code for LOTUS and GAMAOD for researchers to use and reproduce our experiments. Our tools can be extended with new datasets and algorithms. 
\end{itemize}

\section{Background}
This section describes related work regarding Automated Machine learning for unsupervised outlier detection, optimal transport and meta-learning.
\subsection{AutoML for Outlier detection}
AutoML ~\citep{Hutter2019AutomatedML} for unsupervised outlier detection is an extremely hard problem due lack of an optimization metric to perform algorithm selection. One can argue that the use of internal metrics like Excess-Mass ~\citep{emmv}, Mass-Volume ~\citep{emmv} and IREOS ~\citep{IREOS} can make algorithm selection possible. \cite{internalstrategysurvey} shows in their experiments that these internal metrics are computationally very expensive and do not scale well for large datasets. This makes it unfeasible to use these metrics in AutoML tools for most real world scenarios.
\par 
There has been recent research on AutoML for outlier detection. PyODDS and MetaOD ~\citep{Li2020PyODDSAE, metaod} are among the few tools which have been shown to automate outlier detection. 
\par
To the best of our knowledge MetaOD ~\citep{metaod} is the current state of the art meta-learner for model selection on outlier detection tasks for tabular data. MetaOD uses meta-learning as a recommendation engine using landmark meta-features and model based meta-features with collaborative filtering ~\citep{stern2010collaborative} to perform model selection for a given task.
\subsection{Meta learning}
Meta-learning or "learning to learn" in AutoML ~\citep{vanschoren_meta_2019} is the study of learning from historical performances of machine learning models on a variety of tasks and using this knowledge to find better models for new tasks. Meta-learning can help to speed up the model selection process and find better architectures. Meta-learning is often proposed as a solution to \textit{cold start problem}, by initializing the  hyperparameters or search space for the AutoML algorithm. This is often called \textit{warm-starting} for AutoML.

\par
\textbf{Meta-learning in existing AutoML tools:}
Different AutoML tools use different meta-learning schemes to solve this cold start problem. AutoSklearn-2.0 ~\citep{feurer-arxiv20a} learns pipeline portfolios, MetaOD ~\citep{metaod} trains a collaborative filtering based algorithm ~\citep{stern2010collaborative} with landmark-based and model-based metafeatures ~\citep{meta-data}, FLAML uses in-built meta-learned defaults for warm starting. MetaBu ~\citep{metabu} uses Fused Gromov Wasserstein with proximal gradient method on landmark meta-features for warm-starting AutoSklearn(More discussion about LOTUS vs MetaBu is provided in the section \ref{metabuvsLOTUS}). 
\subsection{Optimal transport and dataset distances}
Optimal transport(OT) theory deals with the problem of finding an optimal transport map between two probability measures, often on different metric spaces. It is closely related to Monge's problem ~\citep{Villani2008OptimalTO}, in which one searches for the optimal transport map between two given measures.

An Optimal transport problem consists of minimizing the cost of transporting mass from one distribution to another. For cost function(ground metric) between pair of points, we calculate the cost matrix $C$ with dimensionality $n \times m $, the OT problem minimizes the loss function $L_c(P):=\langle C,P\rangle$ w.r.t a coupling matrix $P$. Most common approach with practitioners is to use a regularized approach which is computationally more efficient $L_c^\epsilon(P):=\langle C,P\rangle + \epsilon r(P)$ where r is negative entropy sinkhorn algorithm \citep{Cuturi2013SinkhornDL} which is computationally more efficient. A discrete OT problem can be defined with two finite pointclouds, 
$\{x^{(i)}\}^{n}_{i=1}$ ,$\{y^{(j)}\}^{m}_{j=1}, x^{(i)},y^{(j)}\in \mathbb{R}^d $, which can be
described as two empirical distributions: $\mu:=\sum^n_{i=1}a_i\delta_{x^{(i)}}, \nu:=\sum^m_{j=1}b_j\delta_{y^{(j)}}$. 
Here $a$ and $b$ are the probability vectors of size $n$ and $m$. 
In this work we are interested in the the Gromov Wasserstein(GW) distance between these two discrete probability distributions. Gromomv Wasserstein allows us to match points taken within different metric spaces. This problem can be written as a function of $(a,A), (b,B)$  \citep{Villani2008OptimalTO, gwlr}:
\begin{equation}
        GW((a,A),(b,B)) = \min_{P\in \Pi_{a,b}} \mathcal{Q}_{A,B}(P)
\end{equation}

Where $\Pi_{a,b}:=\{ P \in \mathbb{R}^{n \times m}_+| P\mathbf{1}_m = a, P^{T}\mathbf{1}_n=b\}$

the energy $\mathcal{Q}_{A,B}$ is a quadratic function of $P$  which can be described as 
\begin{equation}
    \mathcal{Q}_{A,B}(P):= \sum_{i,j,i^{'},j^{'}}(A_{i,i'}-B_{j,j'})^2P_{i,j}P_{i',j'}
\end{equation} 
In this work we are interested in the Entropic Gromov Wasserstein cost ~\citep{pmlr-v48-peyre16}:

\begin{equation}\label{eq 2-1}
    GW_\varepsilon((a,A),(b,B)) = \min_{P\in \Pi_{a,b}} \mathcal{Q}_{A,B}(P) - _\varepsilon H(P)
\end{equation}

where $GW_\epsilon$ is the Entropic Gromov Wasserstein cost between our distributions $A$ and $B$,
and $\varepsilon H(P)$ is the shannon entropy. The problem with Gromov Wasserstein is that it is NP-hard and the entropic approximation of GW still has cubic complexity. To speed up the computations and use it in a realistic AutoML settings we use the Low-Rank Gromov Wasserstein (GW-LR) approximation ~\citep{pmlr-v139-scetbon21a, Scetbon2022LowrankOT, gwlr}, which reduces the computational cost from cubic to linear time. \cite{gwlr} consider the GW problem with low-rank couplings, linked by a common marginal $g$. Therefore, the set of possible transport plans is restricted to those adopting the factorization of the form $P_r = Qdiag(1/g)R^T$. In this form $Q$ and $R$ are thin matrices with dimensionality of $n\times r$, $r\times m$ respectively and $g$ is a $r-$ dimensional probability vector. 
The GW-LR distance is be described as:

\begin{equation}\label{eq 2-2}
    \text{GW-LR}^{(r)}((a,A),(b,B)) := \min_{(Q,R,g)\in \mathcal{C}_{a,b,r}}\mathcal{Q}_{A,B}(Qdiag(1/g)R^T)
\end{equation}

Our primary inspiration for LOTUS comes from two different works. \begin{enumerate}
    \item \citet{AlvarezMelis2020GeometricDD} proposes optimal transport dataset distance(OTDD) which uses optimal transport to learn a mapping over the joint feature and label spaces. \cite{AlvarezMelis2020GeometricDD} proposed that optimal transport distances can be used as a similarity metric between different datasets from different domains and subdomains. 
    \item  Work of \citet{Nies2021TransportDO} argues that optimal transport measures can be used as a correlation measure between two random variables via transport dependency.
\end{enumerate}

 There have been other studies exploring the space of dataset and task similarity with distance measures. \cite{pmlr-v139-gao21a} proposes “coupled transfer distance" which utilises optimal transport distances as a transfer learning distance metric. \cite{10.1093/imaiai/iaaa033} explores connections between Deep Learning, Complexity Theory, and Information Theory through their proposed asymmetric distance on tasks. 

\section{Methodology (Meta learning for unsupervised outlier detection)}

\subsection{problem statement}
In this section, we formally describe the problem of model selection for unsupervised outlier detection. \newline\textbf{Problem Statement:} Given a new dataset without any labels, our meta-learner needs to selects an optimal algorithm with associated hyperparameters from a collection of previously evaluated pipeline. In this setting, we cannot optimize the given model for the dataset as there are no given labels. This problem becomes from a Combined model selection and hyperparameter optimization problem to a \textit{zero-shot model reccomendation problem}.
\newline
Given a new input dataset (i.e., detection task) $D_{new} = ( X_{new} )$ without any labels, Select a model $A^*_{\lambda^*} \in \mathcal{A}$ to employ on $X_{new}$. Where $A^*_{\lambda^*} $ is a tuned model for a similar dataset to $X_{new}$.

\subsection{LOTUS+GAMAOD: A combined framework for automated unsupervised outlier detection}

In this work, we present a meta-learner for outlier detection (LOTUS) and a tool to collect meta-data for automated supervised outlier detection (GAMAOD). We call the training phase of GAMAOD meta-training, where different datasets are supplied to GAMAOD with labels. 
This meta-data contains two major attributes:
\begin{itemize}
    \item A collection of $n$ meta datasets $\mathcal{D}_{meta} = \{D_1, ..., D_n\}$ with labels, test and train splits such that $D_i = (X^{train}_i, y^{train}),(X^{test}, y^{test})$
    \item A collection of $n$ optimized algorithm(s) with associated hyperparamters for every dataset in $\mathcal{D}_{meta}$; $\mathcal{A} = \{A^*_{\lambda^*_1}, ..., A^*_{\lambda^*_n}\}$
\end{itemize}

\begin{figure}[h]
\begin{center}
\includegraphics[width=\textwidth,height=\textheight,keepaspectratio]{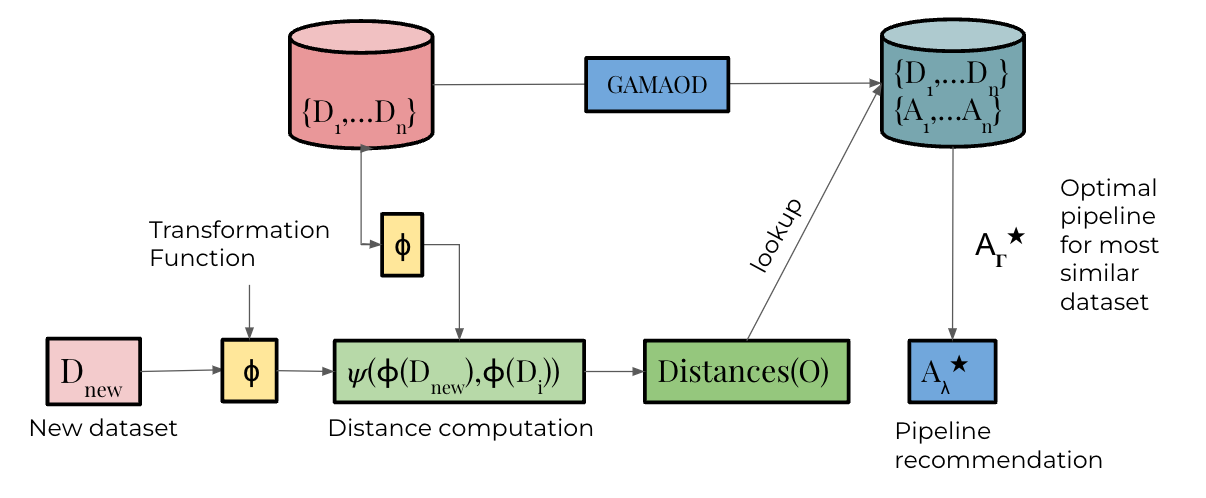}
\end{center}
\caption{An overview of LOTUS}
    \label{fig:LOTUSfig}
\end{figure}

An overview of our system can be found in Figure \ref{fig:LOTUSfig}. We will discuss the building of our meta-data before describing our meta-learning approach.
\subsubsection{GAMAOD: Automated supervised learning for outlier detection}
\textbf{Problem Formulation: }A Combined model selection and hyperparameter optimization problem ~\citep{Thornton2013} for a supervised learning task is as follows:
\begin{equation}
\begin{split}
A^*_{\lambda^*} =
\operatorname*{argmin}_{%
       \substack{%
         \forall A^j \in \boldsymbol{A} \\
         \forall \lambda \in \boldsymbol{\Lambda_A}
       }
     }
\frac{1}{k} 
\sum_{f=1}^{k} L \left( A^j_\lambda, \big\{\boldsymbol{X}^f_{train}, \boldsymbol{y}^f_{train}\big\}, \big\{\boldsymbol{X}^f_{val}, \boldsymbol{y}^f_{val}\big\} \right)
\end{split}
\label{eq:1}
\end{equation}
in equation \ref{eq:1}, $A_{\lambda^*}^*$ is an optimal combination of learning algorithm from search space $A$ with associated hyperparameter space $\Lambda_A$ over $k$ cross validation folds of dataset $D$ where $D=\{X,y\}$ with training and validation splits. $L$ is our evaluation measure.
\par
The CASH problem from equation \ref{eq:1} relies on the validation split to optimise for the optimal configuration. However, in unsupervised outlier detection scenario the learning algorithm does not have access to labels but the AutoML framework does. We cannot do cross validation folds in the unsupervised outlier model selection setting as the learning algorithms are trained without labels and evaluated with labels. Hence performing k-fold CV is not useful in this setting. Our modified CASH formulation to select the optimal unsupervised algorithm with access to labels is as follows:

\begin{equation}
\begin{split}
A^*_{\lambda^*} =
\operatorname*{argmin}_{%
       \substack{%
         \forall A^j \in \boldsymbol{A} \\
         \forall \lambda \in \boldsymbol{\Lambda_A}
       }
     }
 L \left( A^j_\lambda, \big\{\boldsymbol{X}_{train}\} \big\{\boldsymbol{y}_{train}\big\} \right)
\end{split}
\label{eq: 2}
\end{equation}
\par
\begin{figure}
\begin{center}
    \includegraphics[scale=0.35]{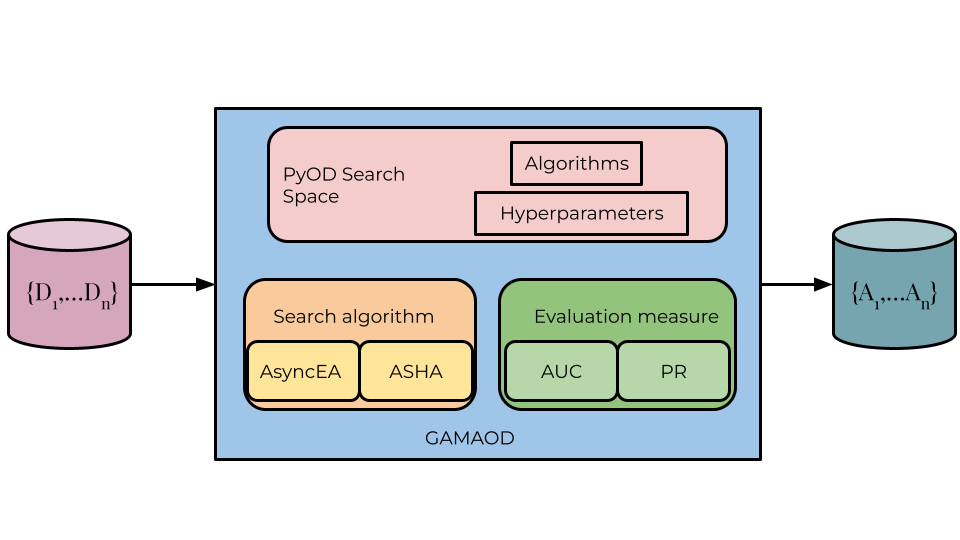}
\caption{An overview of GAMAOD}
    \label{fig:gamaOD}
\end{center}

\end{figure}
\textbf{GAMAOD: }To populate our meta-data we develop an extension on top of GAMA ~\citep{gama}, we call this extension GAMAOD. GAMAOD's search space consists of outlier detectors from PyOD ~\citep{Zhao2019PyODAP}, PyOD is a Python library for detecting outlying objects in multivariate data. GAMAOD can use different metrics to optimise for the given task like AUC score and PR.

\par

\subsubsection{LOTUS: Learning to learn with optimal transport for unsupervised scenarios}
\par We describe LOTUS approach in algorithm \ref{algo: LOTUS}.
 Our meta-learning approach has two components. First, we have a transformation function that is applied to the dataset. Let's call this function $\phi$, which is applied on given datasets $D_a$ and $D_b$. In the second phase, we calculate the dataset similarity $\mathcal{O}$ based on some distance metric $\psi$ in equation \ref{eq: 3-1}. 
Because our distributions are lie on different metric spaces we calculate the Low Rank Gromov Wasserstein distance from equation \ref{eq 2-2} on these transformed distributions in equation \ref{eq: 3-2}. 
\begin{equation}
    \mathcal{O}= \psi(\phi(D_a)\phi(D_b))
    \label{eq: 3-1}
\end{equation}
\begin{equation}
            \mathcal{O}=  \text{GW-LR}^{(r)}(\phi(D_{a})\phi(D_b))
    \label{eq: 3-2}
    \end{equation}
To find the most similar dataset $\Gamma$, we take the smallest distance from datasets from our distance matrix $\{\mathcal{O}_1,...,\mathcal{O}_n\}$ between a new dataset $D_{new}$ and existing datasets from $\mathcal{D}_{meta}$.
which results in: 
\begin{equation} 
    \Gamma = argmin\{\mathcal{O}_1,...,\mathcal{O}_n\}
    \label{eq: 3-3}
\end{equation}
Our meta-approach for capturing the similarity metric between datasets can be described as: 
\begin{equation}
    \Gamma_{\Gamma \in \mathcal{D}_{meta}} = argmin (\text{GW-LR}^{(r)}(\phi(D_{new})\iota\{(\phi(D_{D\in \mathcal{D}_{meta}}))\}))
    \label{eq: 3-4}
\end{equation}
where $\iota$ is the iterator over $\mathcal{D}_{meta}$.
\newline
LOTUS then accesses the optimal pipeline from $\mathcal{A}$: $A^*_{\lambda^*new} = A^*_{\lambda^*\Gamma}$ Where $A^*_{\lambda^*new}$ is the optimal pipeline for $D_{new}$.
\begin{algorithm}
\caption{Pseudocode for LOTUS}\label{algo: LOTUS}
\textbf{Inputs:} \text{$D_{new}, \mathcal{D}_{meta}, \mathcal{A}$}
\begin{algorithmic}
\While{$D_{i} \in \mathcal{D}_{meta}$}
\State \text{$\mathcal{O}_i \gets \psi(\phi(D_{new}, D_i))$} \Comment{Distance calculation}
\EndWhile
\State \text{$\Gamma \gets \mathcal{D}_{meta}[argmin\{\mathcal{O}_1,...,\mathcal{O}_n\}]$} \Comment{Retrieval of most similar dataset}
\State \text{$A^*_{\lambda^*new} \gets A^*_{\lambda^*\Gamma}$} \Comment{Model Selection}
\end{algorithmic}
\end{algorithm}

\section{experiments on ADBench}
For our experiments, we use ADBench ~\citep{Han2022ADBenchAD} and retrieve all tabular datasets. This collection consists of 46 datasets. As we do not have access to multiple benchmarks we use the N-1 strategy for the evaluation of our system, i.e., we take one dataset at a time from ADBench and use the other datasets in the meta-data. This ensures independent meta-training on the following datasets. We compare our approach with 7 outlier detection algorithms available in PyOD ~\citep{Zhao2019PyODAP} and the current state of the art meta-learner for outlier detection MetaOD ~\citep{metaod}.
From PyOD we compare our approach with the following algorithms: IForest ~\citep{Liu2008IsolationF}, ABOD ~\citep{ABOD}, OCSVM ~\citep{ocsvm}, LODA ~\citep{Pevn2015LodaLO}, KNN ~\citep{knn2, knn1}, HBOS ~\citep{hbos}.

\par For experimental consistency, we use the same search space for GAMAOD pretraining as MetaOD (\ref{metaodrepro}) to ensure a fair comparison. To populate our meta-data, we run GAMAOD for 2 hours on every ADBench dataset to find the optimal pipeline for a given dataset. We use an asynchronous evolutionary algorithm to iterate over the search space and return the optimal pipeline. GAMAOD uses AUC score as internal metric for model selection. 

\textbf{Implementation details: }We use Independent Component Analysis(ICA) ~\citep{Hyvrinen2000IndependentCA} from scikit-learn ~\citep{scikit-learn} as our transformation function $\phi$. We use OTT-JAX library ~\citep{Cuturi2022OptimalTT} library to implement Low Rank Gromov Wassersstein distance. For this experiment, we set the rank parameter of Low Rank Gromov Wasserstein to 6. The model selection phase of LOTUS in our experiments is as follows:
First the datasets are transformed via ICA and then converted into JAX ~\citep{jax2018github} pointclouds geometry objects \footnote{\url{https://ott-jax.readthedocs.io/en/latest/\_autosummary/ott.geometry.pointcloud.PointCloud.html}} and then we turn these distributions into a quadratic regularized optimal transport problem ~\citep{pmlr-v48-peyre16}. We input this quadratic problem to our Gromov Wasserstein Low Rank solver which returns us the distance(cost) between two datasets. When a new dataset is given to LOTUS, the pipeline corresponding to the dataset with the lowest distance(except the new dataset itself) is chosen from the optimal pipeline database. 

\section{results and discussion}
\subsection{Experimental results}
We use the Bayesian Wilcoxon signed-rank test (or ROPE test, ~\cite{Ropetutorial, rope2}) to analyze the results of our experiments. ROPE defines an interval wherein the differences in model performance are considered equivalent to the null value. Using this test allows us to compare model performances in a more practical sense. We set the ROPE value to 1\% for our experiments. We use the baycomp library ~\citep{Ropetutorial} to run and visualize the analyses.



\par
\begin{figure}[h]
\begin{center}
\includegraphics[scale=0.5]{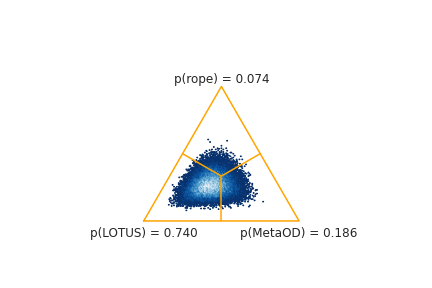}
\end{center}
\caption{ROPE test LOTUS vs MetaOD.}
    \label{fig:rope}
\end{figure}

\textbf{LOTUS vs MetaOD:} The ROPE test on AUC scores comparing LOTUS and MetaOD are shown in Figure \ref{fig:rope}. There is a 74.2 \% probability that LOTUS is better than MetaOD. LOTUS proves to be more robust than MetaOD, since $p(LOTUS) > p(MetaOD)$. We show the per-dataset performances in Appendix \ref{scores}.

\par
\begin{figure}
     \centering
     \begin{subfigure}[b]{0.3\textwidth}
         \centering
         \includegraphics[width=\textwidth]{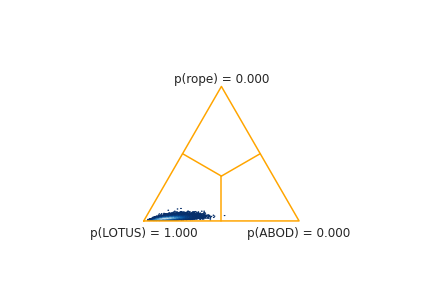}
         \caption{LOTUS vs ABOD}
         \label{fig:y equals x}
     \end{subfigure}
     \hfill
     \begin{subfigure}[b]{0.3\textwidth}
         \centering
         \includegraphics[width=\textwidth]{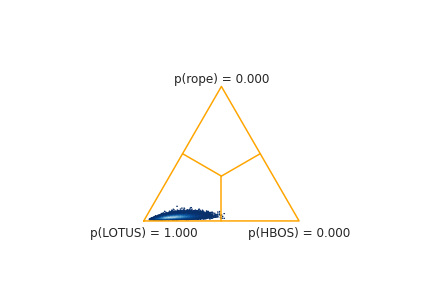}
         \caption{LOTUS vs HBOS}
         \label{fig:three sin x}
     \end{subfigure}
     \hfill
     \begin{subfigure}[b]{0.3\textwidth}
         \centering
         \includegraphics[width=\textwidth]{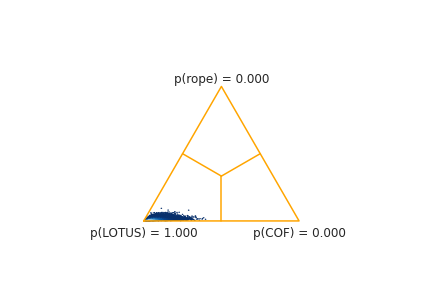}
         \caption{LOTUS vs COF}
         \label{fig:five over x}
     \end{subfigure}
        \label{fig:three graphs}
      \centering
     \begin{subfigure}[b]{0.3\textwidth}
         \centering
         \includegraphics[width=\textwidth]{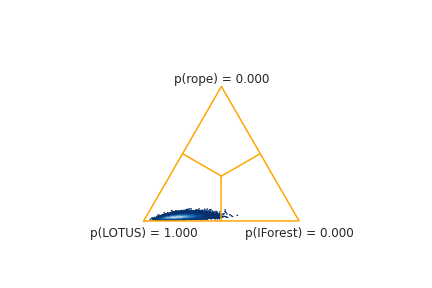}
         \caption{LOTUS vs IForest}
         \label{fig:y equals x}
     \end{subfigure}
     \hfill
     \begin{subfigure}[b]{0.3\textwidth}
         \centering
         \includegraphics[width=\textwidth]{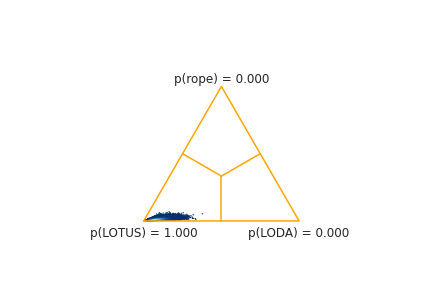}
         \caption{LOTUS vs LODA}
         \label{fig:three sin x}
     \end{subfigure}
     \hfill
     \begin{subfigure}[b]{0.3\textwidth}
         \centering
         \includegraphics[width=\textwidth]{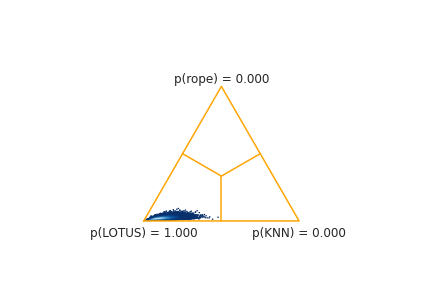}
         \caption{LOTUS vs KNN}
         \label{fig:five over x}
     \end{subfigure}
    \hfill
     \begin{subfigure}[b]{0.3\textwidth}
         \centering
         \includegraphics[width=\textwidth]{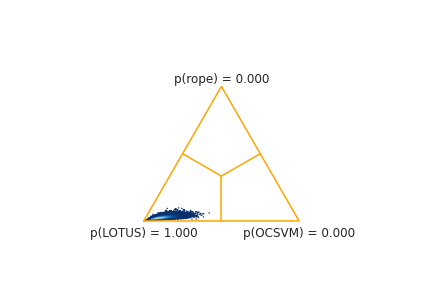}
         \caption{LOTUS vs OCSVM}
         \label{fig:five over x}
     \end{subfigure}
        \caption{ROPE test result of LOTUS vs (a) ABOD (b) HBOS (c) COF (d) IForest (e) LODA (f) KNN (g) OCSVM}
        \label{fig: all}       
\end{figure}

\textbf{Results against other estimators:} ROPE testing on the results of ROCAUC performance of LOTUS vs other estimators are shown in Table \ref{tab: ropeall}. LOTUS proves to be significantly better than other techniques, with default parameters, in PyOD. In this case $P(LOTUS) >> P(PyOD_{estimators})$, Figure \ref{fig: all} shows the simplex plots of MetaOD vs other estimators. We also include a Critical Difference Plot of performances between LOTUS and PyOD estimators (lower is better) in Figure \ref{fig:avgrank}. The detailed experimental results are reported in appendix \ref{scores} table \ref{tab: LOTUSvsall}.
\begin{figure}
    \centering
    \includegraphics[ scale=0.5]{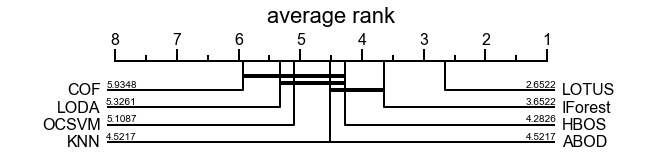}
    \caption{Comparison of average rank
(lower is better) of methods w.r.t. performance across datasets in ADBench.}
    \label{fig:avgrank}
\end{figure}
\begin{table}[]
    \centering
        \begin{tabular}{lrrr}
        \hline Estimator name &   p(LOTUS) &  p(rope) &  p(Estimator) \\
        \hline       IForest &  0.99954 &   0.0 &    0.00046 \\
        ABOD &  1.0 &   0.0 &    0.0\\
        OCSVM &  1.0 &   0.0 &    0.0 \\
        LODA &  1.0 &   0.0 &    0.0\\  
        KNN &  1.0 &   0.0 &    0.0 \\  
        HBOS &  0.99982 &   0.0 &    0.00018 \\ 
        COF &  1.0 &   0.0 &    0.0 \\
        \hline\end{tabular}
    \caption{Rope testing results with LOTUS vs PyOD estimators with rope=1\%}
    \label{tab: ropeall}
\end{table}
\subsection{Using optimal transport distances as a similarity measure}

In our experiments, we show that LOTUS is more robust and better than current state of the art meta-learner  MetaOD for unsupervised outlier detection tasks and other outlier detection algorithms in default configuration.
\par In our method we experimentally show that using optimal transport distances like GW-LR is a feasible approach for dataset similarity and meta-learning. We would like to emphasize that this similarity measure should only be used as a relative similarity measure, for e.g. in our case where we use this similarity measure to find the most similar dataset from a collection of datasets in $\mathcal{D}_{meta}$. To estimate to what degree datasets are similar \cite{Nies2021TransportDO} proposes optimal transport based correlation measures that can be leveraged. Our approach assumes that Wasserstein distances can capture intrinsic properties of datasets and can capture the similarity between them, \cite{AlvarezMelis2020GeometricDD} also proposes their approach with optimal transport distances to provide some sort of distance between dataset.

\subsection{Related Works}
In this section we will discuss the difference between closest approaches to LOTUS which are MetaOD and MetaBu.
\subsubsection{LOTUS vs MetaOD}
LOTUS and MetaOD solve the same problem of model selection problem for unsupervised outlier detection. The major difference in LOTUS and MetaOD is meta-feature generation. LOTUS aims to capture the similarity of the given source and target representations via optimal transport. MetaOD captures similarity with a combination of landmark-features and model-based features and uses a rank-based criteria called discounted cumulative gain for model selection. MetaOD also uses stochastic algorithms such as Isolation Forest and LODA for model-based meta-feature generation which means that the absolute dataset similarity and ranking can differ based on the number of runs. LOTUS on the other hand uses non stochastic methods for distance calculation. Our approach is generalises more than MetaOD as well for different unsupervised tasks as it aims to find similar dataset independent of task, whereas MetaOD's similarity is highly coupled with the task of outlier detection.
\subsubsection{LOTUS vs MetaBu}\label{metabuvsLOTUS}
 MetaBu ~\citep{metabu} was proposed as a solution to cold start problem in supervised learning scenario. \cite{metabu} uses Fused-Gromov-Wasserstein distance with multi dimensional scaling ~\citep{Cox2008} by first extracting meta-features from the target representation and source representation and proximal gradient method ~\citep{proximalgrad}. LOTUS on the other hand uses independent component analysis and Low Rank Gromov Wasserstein. LOTUS is much faster than MetaBu simply because of computational complexity differences between Low Rank Gromov Wasserstein($O(n)$) and Fused Gromov Wasserstein($O(n^3)$). LOTUS is a simpler approach as well as compared to MetaBu(2 phases vs 5 phases). Lastly, LOTUS is a solution for unsupervised setting whereas MetaBu relies on landmark features from PyMFE ~\citep{PyMFE} which are more reliable for datasets with labels. Similar to MetaOD, MetaBu setting is limited to only one task (supervised classification) as it relies on landmark-features which require labels. LOTUS therefore can generalize to more tasks than MetaBu.
 
\subsection{limitations}
\begin{enumerate}
    \item \textbf{LOTUS}: LOTUS depends on the quality of meta-data, i.e. range of datasets and algorithms in our case. In the worst case scenario, if there are no similar datasets in the $\mathcal{D}_{meta}$, LOTUS can recommend a dataset which is not sufficiently similar to new dataset. On the other hand, it is expected to improve as more benchmarks and datasets with different properties become available. 
    \item \textbf{OT distance:} The computation cost of GW-LR on really large datasets can still be very high. In these cases we recommend using stratified sampling or random sampling depending on the nature of dataset and problem. 
    \item \textbf{Parameter tuning:} Tuning rank of GW-LR can be tricky. Low rank can result in faster computation but high loss and high rank can result in less efficient algorithm. \cite{gwlr} states an experiment where they study the affect of rank of GW-LR. This rank can also be tuned by minimizing the loss between GW and GW-LR. 
\end{enumerate}

\section{conclusion and future work}
Model selection for unsupervised outlier detection is a challenging task. We do not have efficient internal metrics for evaluating an algorithm without ground truth.  In this work, we proposed a new meta-learner: \textbf{LOTUS}, which uses optimal transport distances to capture the similarity between datasets and uses that similarity measure to recommend pipelines from a meta-data. To run these experiments, we also developed another tool \textbf{GAMAOD} which is an extension of GAMA and allows users to find optimal outlier detection algorithms in a supervised setting. Through our experiments, we demonstrate that LOTUS outperforms MetaOD and other built-in estimators in PyOD. The LOTUS approach also enables researchers to use a simplified meta-learning framework as compared to other landmark and model-based meta-features methods where meta-features are highly specialized according to the domain.
\par
 We believe that our approach can be easily extended as a meta-learner to perform model selection in other unsupervised machine learning tasks as well. These include clustering, distance metric learning, density estimation and covariance estimation. This approach can also be used as a meta-learner to warm-start neural architecture search(NAS) problems. 

\par



\section{reproducibilty statement}
We opensource both LOTUS and GAMAOD with hyperparameters used for this experiment. We also provide scripts which can be used to perform these experiment for just one dataset without making the meta-data first(not reccomended). We aim to provide modularity to researchers therefore we users them to save and retrieve meta-data in whatever format they want. More information about reproducing our experiments can be found in the README.md of the supplementary code repository. To reproduce LOTUS for other tasks and dataset, users are simply required to change the datasets and algorithms in meta-data. The approach works out of the box for other scenarios. While reproducing the experiments, the results can differ due to stochasticity of few algorithms. 
\bibliography{iclr2023_conference}
\bibliographystyle{iclr2023_conference}

\newpage
\appendix
\section{Appendix}
\subsection{Performances}\label{scores}
Table \ref{tab: MetaodvsLOTUS score} contains the performances of LOTUS and MetaOD on 42 datasets, We had to eliminate 4 datasets from this experiment because MetaOD returned invalid models for these datasets(i.e. models with invalid values). Scores are in bold where AUC of LOTUS $>$ MetaOD or differ by less than a \%. The dataset names are as they were in ADBench ~\citep{Han2022ADBenchAD}.
\newline
\begin{table}[]
    \centering
\begin{tabular}{|c|c|c|}\hline{}      Dataset &  LOTUS &  MetaOD\\ \hline
19\_landsat &           \textbf{0.7902} &         0.5931 \\
25\_musk &           \textbf{0.9895} &         0.9655 \\
24\_mnist &           \textbf{1.0000} &         1.0000 \\
32\_shuttle &           \textbf{0.9216} &         \textbf{0.9163} \\
23\_mammography &           \textbf{0.6434} &         0.6477 \\
42\_WBC &           0.8521 &         \textbf{0.8655} \\
15\_Hepatitis &           \textbf{0.9353} &        \textbf{ 0.9353} \\
43\_WDBC &           0.8548 &         \textbf{0.9671} \\
12\_fault &           \textbf{0.9246} &         0.9043 \\
10\_cover &           \textbf{0.9463} &         \textbf{0.9436} \\
34\_smtp &           0.2744 &         \textbf{0.5212} \\
11\_donors &           \textbf{0.8064} &         \textbf{0.8049} \\
29\_Pima &           \textbf{0.8804} &         0.7197 \\
37\_Stamps &           \textbf{0.9275} &         \textbf{0.9339} \\
44\_Wilt &           \textbf{0.7765} &         0.5327 \\
40\_vowels &           0.8491 &         \textbf{0.9355} \\
8\_celeba &           \textbf{0.9908} &         0.9906 \\
1\_ALOI &           \textbf{0.8954} &         0.8957 \\
30\_satellite &           \textbf{0.8913} &         0.7890 \\
26\_optdigits &           \textbf{0.9996} &         0.9997 \\
2\_annthyroid &          \textbf{ 0.8472} &         0.8445 \\
41\_Waveform &           \textbf{0.9758} &         0.9413 \\
28\_pendigits &           0.8597 &         \textbf{0.9265} \\
4\_breastw &           \textbf{0.7466} &         0.7438 \\
21\_Lymphography &           0.9441 &         \textbf{0.9861} \\
20\_letter &           0.9701 &         \textbf{0.9891} \\
39\_vertebral &           0.7634 &         \textbf{0.8424} \\
47\_yeast &           \textbf{0.9089} &         \textbf{0.9097} \\
3\_backdoor &           \textbf{1.0000} &         1.0000 \\
13\_fraud &           \textbf{0.9646} &         0.8904 \\
45\_wine &           \textbf{0.9841} &         0.9481 \\
22\_magic.gamma &          \textbf{ 0.9322} &         0.8122 \\
9\_census &           0.9819 &         \textbf{1.0000} \\
7\_Cardiotocography &           \textbf{0.9392} &         \textbf{0.9378} \\
35\_SpamBase &           \textbf{0.9446} &         0.9015 \\
46\_WPBC &           0.7811 &        \textbf{ 0.8088 }\\
36\_speech &           \textbf{1.0000} &         0.4344 \\
6\_cardio &           \textbf{0.9794} &         \textbf{0.9793} \\
31\_satimage-2 &           \textbf{0.9552} &         0.8100 \\
18\_Ionosphere &           0.8072 &         \textbf{0.8338} \\
27\_PageBlocks &           0.7164 &         \textbf{0.7668} \\
          5\_campaign &           \textbf{0.9922} &         \textbf{0.9996} \\
          \hline
          \end{tabular}
\caption{AUC scores of MetaOD vs LOTUS on ADBench}
    \label{tab: MetaodvsLOTUS score}
\end{table}

Table \ref{tab: LOTUSvsall} reports the auc scores over datasets from ADBench. The bold number shows scores where LOTUS is better than \textbf{all} other estimators in PyOD.
\begin{table}[]
    \centering
\begin{tabular}{|c|c|c|c|c|c|c|c|c|c|}
\hline
\textbf{Dataset} &   \textbf{IForest} &      \textbf{ABOD} &     \textbf{OCSVM} &      \textbf{LODA} &       \textbf{KNN} &   \textbf{HBOS} &       \textbf{COF} &  \textbf{LOTUS} \\ \hline           
44\_Wilt &  0.471963 &  0.568222 &  0.301310 &  0.408280 &  0.472095 &  0.281412 &  0.544269 &         \textbf{0.7765} \\           
6\_cardio &  0.943738 &  0.498576 &  0.939676 &  0.892753 &  0.741544 &  0.865343 &  0.544550 &         \textbf{0.9794} \\            
43\_WDBC &  0.987241 &  0.987241 &  0.989655 &  0.987586 &  0.960345 &  \textbf{0.998966} &  0.771034 &         0.8548 \\          
4\_breastw &  0.976321 &  0.976321 &  0.778694 &  \textbf{0.981964} &  0.947386 &  0.969329 &  0.381366 &         0.7466 \\            
42\_WBC &  \textbf{0.993567} &  \textbf{0.993567} &  \textbf{0.994103} &  \textbf{0.995980} &  0.911954 &  \textbf{0.991691} &  0.754757 &         0.8521 \\           
47\_yeast &  0.431011 &  0.417114 &  0.448353 &  0.492504 &  0.413668 &  0.410032 &  0.428639 &         \textbf{0.9089} \\            
45\_wine &  0.735205 &  0.735205 &  0.681612 &  0.923158 &  0.471241 &  0.891757 &  0.412289 &         \textbf{0.9841} \\         
5\_campaign &  0.692549 &  0.642977 &  0.645556 &  0.566477 &  0.696817 &  0.771387 &  0.564588 &         \textbf{0.9922} \\            
46\_WPBC &  0.522489 &  0.522489 &  0.475911 &  0.562133 &  0.419170 &  0.555259 &  0.495170 &         \textbf{0.7811} \\ 
7\_Cardiotocography &  0.752439 &  0.539423 &  0.810433 &  0.785916 &  0.582569 &  0.623355 &  0.572511 &         \textbf{0.9392} \\ 
8\_celeba &  0.757810 &  0.757810 &  0.761861 &  0.718291 &  0.632204 &  0.805965 &  0.393545 &         \textbf{0.9908} \\           
9\_census &  0.598140 &  0.598140 &  0.523211 &  0.325589 &  0.650628 &  0.633393 &  0.413254 &        \textbf{ 0.9819} \\       
39\_vertebral &  0.377788 &  0.377788 &  0.427308 &  0.284423 &  0.417163 &  0.282356 &  0.321923 &         \textbf{0.7634} \\       
41\_Waveform &  0.669757 &  0.698172 &  0.474443 &  0.611266 &  0.782120 &  0.639714 &  0.804121 &        \textbf{ 0.9758} \\        
38\_thyroid &  \textbf{0.979620} &  \textbf{0.979620 }&  0.867786 &  0.699534 &  0.951152 &  0.952834 &  0.871991 &         0.7910 \\          
40\_vowels &  0.708373 &  0.956714 &  0.532701 &  0.655924 &  \textbf{0.971722} &  0.646130 &  0.849763 &         0.8491 \\         
3\_backdoor &  0.734361 &  0.734361 &  0.802264 &  0.708914 &  0.738679 &  0.665487 &  0.728995 &         \textbf{1.0000} \\        
32\_shuttle &  0.996250 &  0.618768 &  0.987461 &  0.951075 &  0.678578 &  0.994925 &  0.557606 &         0.9216 \\      
31\_satimage-2 &  0.996844 &  0.762625 &  0.983527 &  0.987126 &  0.909884 &  0.985936 &  0.451384 &         0.9552 \\       
26\_optdigits &  0.771433 &  0.525541 &  0.527237 &  0.623480 &  0.398194 &  0.852822 &  0.423611 &         \textbf{0.9996} \\            
1\_ALOI &  0.501898 &  0.609567 &  0.532848 &  0.549594 &  0.555634 &  0.478001 &  0.635583 &         \textbf{0.8954} \\        
35\_SpamBase &  0.657074 &  0.390792 &  0.520510 &  0.273952 &  0.515358 &  0.651507 &  0.416468 &         \textbf{0.9446} \\          
36\_speech &  0.469975 &  0.729473 &  0.462061 &  0.448529 &  0.473192 &  0.476358 &  0.553156 &         \textbf{1.0000} \\            
34\_smtp &  0.696899 &  0.670223 &  0.018006 &  0.372124 &  0.744582 &  0.878626 &  0.890630 &         0.2744 \\     
22\_magic.gamma &  0.704407 &  0.799144 &  0.594241 &  0.635940 &  0.823228 &  0.681717 &  0.663549 &         \textbf{0.9322} \\     23\_mammography &  0.859409 &  0.859409 &  0.854704 &  0.814810 &  0.859614 &  0.871755 &  0.792004 &         0.6434 \\           24\_mnist &  0.794443 &  0.750330 &  0.834765 &  0.743575 &  0.828259 &  0.619057 &  0.733384 &        \textbf{ 1.0000} \\          
20\_letter &  0.581556 &  0.880889 &  0.485185 &  0.627407 &  0.867111 &  0.540593 &  0.829704 &         \textbf{0.9701} \\       
30\_satellite &  0.707795 &  0.538013 &  0.605468 &  0.609243 &  0.646056 &  0.768130 &  0.556999 &         \textbf{0.8913} \\         
19\_landsat &  0.495534 &  0.500057 &  0.374050 &  0.382382 &  0.577134 &  0.556768 &  0.542057 &         \textbf{0.7902} \\          
37\_Stamps &  0.909527 &  0.909527 &  0.878255 &  0.944582 &  0.746473 &  0.928582 &  0.636364 &         0.9275 \\      
18\_Ionosphere &  0.867847 &  0.867847 &  0.765359 &  0.858325 &  0.862297 &  0.667416 &  0.850478 &         0.8072 \\    
21\_Lymphography &  0.997003 &  0.997003 &  0.993506 &  0.667582 &  0.512862 &  0.995005 &  0.934316 &         0.9441 \\            
25\_musk &  0.999923 &  0.085936 &  0.818675 &  0.959047 &  0.701124 &  1.000000 &  0.400387 &         0.9895 \\     
17\_InternetAds &  0.700473 &  0.673305 &  0.710028 &  0.580881 &  0.712320 &  0.704318 &  0.693902 &         \textbf{1.0000 }\\        
16\_http &  1.000000 &  1.000000 &  0.995308 &  0.000000 &  0.001340 &  0.994638 &  0.583110 &         0.7106 \\       
15\_Hepatitis &  0.742736 &  0.742736 &  0.722262 &  0.772817 &  0.467871 &  0.813292 &  0.425388 &         \textbf{0.9353} \\           
14\_glass &  0.818496 &  0.818496 &  0.459264 &  0.632274 &  0.740799 &  0.791758 &  0.882668 &         0.8374 \\           
13\_fraud &  0.934023 &  0.941569 &  0.914391 &  0.751185 &  0.916394 &  0.941169 &  0.914591 &         \textbf{0.9646} \\          
11\_donors &  0.794215 &  0.794215 &  0.723436 &  0.260784 &  0.829936 &  0.763981 &  0.720262 &         \textbf{0.8064} \\           
12\_fault &  0.571477 &  0.676490 &  0.494426 &  0.436072 &  0.713079 &  0.479224 &  0.612146 &         \textbf{0.9246} \\       
2\_annthyroid &  0.824922 &  0.824922 &  0.606069 &  0.305845 &  0.730291 &  0.691522 &  0.704828 &         \textbf{0.8472 }\\      
27\_PageBlocks &  0.889696 &  0.684494 &  0.892650 &  0.753280 &  0.769997 &  0.788657 &  0.673234 &         0.7164 \\       
28\_pendigits &  0.949714 &  0.673023 &  0.938642 &  0.951140 &  0.705836 &  0.921169 &  0.475639 &         0.8597 \\            
29\_Pima &  0.660016 &  0.660016 &  0.580166 &  0.606169 &  0.685681 &  0.713573 &  0.566752 &         \textbf{0.8804} \\           
10\_cover &  0.914310 &  0.767605 &  0.886407 &  0.866889 &  0.899776 &  0.795243 &  0.870260 &         \textbf{0.9463}\\ \hline\end{tabular}
    \caption{AUC Scores: LOTUS vs PyOD estimators with default configuration}
    \label{tab: LOTUSvsall}
\end{table}

\subsection{Baslines}\label{baselinedefinition}
The 8 baslines estimators and frameworks are listed below with brief description from PyOD's ~\citep{Zhao2019PyODAP} documentation for reference here:
\begin{enumerate}
    \item \textbf{MetaOD}: MetaOD is the first automated tool for outlier detection. MetaOD use collaborative filtering, landmark and model based meta-features to recommend the model for given task.
    \item \textbf{IForest}:  IsolationForest ‘isolates’ observations by randomly selecting a feature and then randomly selecting a split value between the maximum and minimum values of the selected feature. 
    \item \textbf{LOF}:The anomaly score of each sample is called Local Outlier Factor. It measures the local deviation of density of a given sample with respect to its neighbors. It is local in that the anomaly score depends on how isolated the object is with respect to the surrounding neighborhood. More precisely, locality is given by k-nearest neighbors, whose distance is used to estimate the local density. By comparing the local density of a sample to the local densities of its neighbors, one can identify samples that have a substantially lower density than their neighbors. These are considered outliers.
    \item \textbf{ABOD}:For an observation, the variance of its weighted cosine scores to all neighbors could be viewed as the outlying score.
    \item \textbf{HBOS}: Histogram- based outlier detection assumes the feature independence and calculates the degree of outlier by building histograms.
    \item \textbf{KNN}: kNN class for outlier detection. For an observation, its distance to its kth nearest neighbor could be viewed as the outlying score.
    \item \textbf{COF}: Connectivity-Based Outlier Factor uses the ratio of average chaining distance of data point and the average of average chaining distance of k nearest neighbor of the data point, as the outlier score for observations.
    \item \textbf{LDOA}: Lightweight on-line detector of anomalies detects anomalies in a dataset by computing the likelihood of data points using an ensemble of one-dimensional histograms.
    \item \textbf{OCSVM}: One class support vector machines unsupervised outlier Detection. Estimate the support of a high-dimensional distribution.
\end{enumerate}
\subsection{LOTUS+GAMAOD search space and MetaOD reproducibility}\label{metaodrepro}
We implement the same searchspace as MetaOD github repository for a fair comparison. \footnote{\url{https://github.com/yzhao062/MetaOD/blob/master/metaod/models/base\_detectors.py}}, MetaOD also uses all the existing datasets from ADbench. We believe that we have fairly evaluated MetaOD against out baseline. We believe that our Benchmark setting was more challenging than the one evaluated in \cite{metaod} where it take child and parent datasets. \footnote{\url{https://github.com/yzhao062/MetaOD/blob/2a8ed2761468d2f8ee2cd8194ce36b0f817576d1/metaod/models/train_metaod.py}}

\end{document}